\documentclass[journal]{IEEEtran}

\usepackage{times}
\usepackage{epsfig}
\usepackage{graphicx}
\usepackage{amsmath}
\usepackage{amssymb}
\usepackage[linesnumbered,ruled]{algorithm2e}
\usepackage{multirow}
\usepackage{xcolor}
\usepackage[normalem]{ulem}
\usepackage{slashbox}
\usepackage{caption}
\usepackage{subcaption}
\usepackage{dblfloatfix}

\usepackage{tabularx}
\usepackage{array}
\usepackage{enumitem}

\makeatletter
\newcommand{\thickhline}{%
	\noalign {\ifnum 0=`}\fi \hrule height 1pt
	\futurelet \reserved@a \@xhline
}
\newcolumntype{"}{@{\hskip\tabcolsep\vrule width 1pt\hskip\tabcolsep}}
\makeatother

\makeatletter
\newcommand{\nosemic}{\renewcommand{\@endalgocfline}{\relax}}
\newcommand{\dosemic}{\renewcommand{\@endalgocfline}{\algocf@endline}}
\let\oldnl\nl
\newcommand{\nonl}{\renewcommand{\nl}{\let\nl\oldnl}}
\makeatother

\ifCLASSINFOpdf
\else
\fi
\begin{document}
%
\title{Spatial-Spectral Regularized Local Scaling Cut for Dimensionality Reduction in Hyperspectral Image Classification}

\author{Ramanarayan~Mohanty,~\IEEEmembership{Student Member,~IEEE,}
        S~L~Happy,~\IEEEmembership{Student Member,~IEEE,}
        and~Aurobinda~Routray,~\IEEEmembership{Member,~IEEE}
\thanks{R. Mohanty is with the Advanced Technology Development Centre, Indian Institute of Technology, Kharagpur,
West Bengal, 721302 India e-mail: ramanarayan@iitkgp.ac.in.}
\thanks{S L Happy and A. Routray are with Department of Electrical Engineering, Indian Institute of Technology, Kharagpur.}}

\maketitle

\begin{abstract}
Dimensionality reduction (DR) methods have attracted extensive attention to provide discriminative information and reduce the computational burden of hyperspectral image (HSI) classification. However, the DR methods face many challenges due to limited training samples with high dimensional spectra. To address this issue, a graph-based spatial and spectral regularized local scaling cut (SSRLSC) for DR of HSI data is proposed. The underlying idea of the proposed method is to utilize the information from both the spectral and spatial domains to achieve better classification accuracy than its spectral domain counterpart. In SSRLSC, a guided filter is initially used to smoothen and homogenize the pixels of the HSI data in order to preserve the pixel consistency. This is followed by generation of between-class and within-class dissimilarity matrices in both spectral and spatial domains by regularized local scaling cut (RLSC) and neighboring pixel local scaling cut (NPLSC) respectively. Finally, we obtain the projection matrix by optimizing the updated spatial-spectral between-class and total-class dissimilarity. The effectiveness of the proposed DR algorithm is illustrated with two popular real-world HSI datasets. 
\end{abstract}


%
\IEEEpeerreviewmaketitle

\section{Introduction}
 \IEEEPARstart {H}{yperspectral} remote sensing images (HSI) with high spectral and spatial resolution capture the inherent properties of the land cover. The wide range of spectral bands of HSI data carries a wealth of information about the surface. 
Hence, the conventional hyperspectral classification systems solely concentrate on the spectral features of a pixel by ignoring the spatial neighborhood information \cite{zhang2015scaling,zhang2013semisupervised,li2012locality}. 

Although these spectral based methods provide satisfactory performances, they possess several limitations, such as: 1) Relatively large spectral bands with respect to small training samples create a singularity in the sample covariance matrix that leads to ill-posed problems in classification. 2) The high intra-class spectral variability leads to class identification problem (e.g. roofs with shadows) and low inter-class spectral variability leads to class discrimination problem (e.g. roads and roofs are similar in spectral domain) \cite{he2017recent} in high resolution HSI. This implies that only spectral similarity measure is not sufficient for HSI data. 
Hence, the spectral similarity measure without considering the spatial inter-pixel correlation sometimes leads to undesired results during classification \cite{zhou2015dimension}. Therefore, use of spectral information alone results in unsatisfactory performances.

To mitigate the challenge of limited number of training samples, several dimensionality reduction (DR) methods adopt the semi-supervised approach by adding the unlabeled samples during training \cite{luo2016semisupervised,liao2013semisupervised,zhang2013semisupervised}. However, this is not sufficient to cope with other mentioned limitations. Hence, supervised DR methods consider both the spatial as well as spectral information to perform the similarity measure. 
The spatial information boosts the spectral based pixel-wise classification. Therefore, spectral-spatial based methods gain considerable attention in HSI feature selection and classification tasks \cite{luo2018feature,zhou2015dimension,xia2015spectral,sun2015supervised}. Some spatial-spectral methods follow the vector stacking method \cite{zhang2016multidomain} by concatenating different spatial and spectral feature vectors into a long vector, while others incorporate the ensemble features in direct classification by using SVM-based multi classifier model \cite{huang2013svm}. Unfortunately, the high dimensional feature vectors with small training sample size in vector stacking method of \cite{zhang2016multidomain} is prone to the curse of dimensionality and fails to discriminate the consistent and complementary features efficiently. Whereas, in \cite{huang2013svm} the spatial information is incorporated in a post-processing step to boost the classification performance.%

In this paper, we propose a graph-based local scaling cut (LSC) method by using both spectral and spatial information. The LSC \cite{zhang2015scaling} uses the spectral information to explore the intrinsic manifold structure of HSI data by constructing pairwise dissimilarity matrix. However, it ignores the spatial properties as well as it faces data singularity issue due to small size training samples for classification. This motivated us to propose a spectral-spatial local scaling cut method with a regularization in spectral LSC.

In the initial step, the proposed spectral-spatial regularized local scaling cut (SSRLSC) processes the HSI data cube spatially by utilizing guided filter \cite{kang2014spectral} to enhance the pixel consistency or homogeneity. The discriminant spectral information of the filtered data is extracted by the proposed spectral-domain regularized LSC (RLSC). Similarly, the spatial domain segmentation is carried out by the proposed neighboring pixel LSC (NPLSC) method. Finally, both the spatial and spectral features are fused to boost the classification task. The advantage of the proposed method is that it embeds the unlabeled spectral distance weight in determining the spatial neighborhood pixels. Therefore, robustness towards the noise increases and spatial neighborhood pixel labels become consistent. The regularization term in the spectral LSC addresses the limitation of data singularity problem caused by small size samples. The regularizer also increases the class discrimination by enhancing the interclass variability. The spectral RLSC also preserves the original complex distribution and local manifold of the data by embedding a localized $k$-nearest neighbor graph. The NPLSC combine the unlabeled spectral weight with the local graph cut based segmentation strategy on the homogeneous HSI data to maintain the label consistency in the spatial domain. By fusing all these, the SSRLSC preserves both the spectral domain local label neighborhood relation and the spatial domain local pixel neighborhood relation and projects the HSI data to a more discriminative space to achieve better classification accuracy.

\section{LSC Criterion}
In order to exploit the intrinsic geometry and local manifold structure of the data, LSC criterion is formulated using localized K-nn graph \cite{zhang2013semisupervised}.

Suppose there are $N$ training samples of $D$ dimensions $x_i \in R^D$, $i = 1,2, ... , N$, with labels $L_i|_{i=1}^N \in \{1,2,...,C \}$, and $C$ is the number of classes. The input training data set is denoted by $X = \{x_i, L_i\}|_{i=1}^N \in R^{D \times N}$. In order to reveal the intrinsic geometry and manifold structure of the data, LSC constructs a between-class dissimilarity matrix $S_b^{spec}$ and within-class dissimilarity matrix $S_w^{spec}$, given by 
\begin{align}
S_b^{spec} = \sum _{c=1}^C {\sum _{x_i \in U_c}} \sum _{x_j \in K^b(x_i)} N_{ij}^b(x_i-x_j)(x_i-x_j)^T \\
S_w^{spec} = \sum _{c=1}^C {\sum _{x_i \in U_c}} \sum _{x_j \in K^w(x_i)} N_{ij}^w(x_i-x_j)(x_i-x_j)^T
\end{align}
where, $K^b(x_i)$ represents the $k_b$ nearest neighbors of $x_i$ from the dissimilar classes (i.e. $K^b(x_i) \in \bar{U_c}$), and $K^w(x_i)$ represents the $k_w$ nearest neighbors of $x_i$ from the same class (i.e. $K^w(x_i) \in {U_c}$). Here, $U_c$ denotes all the samples in $c$th class and $\bar{U_c}$ denotes all the samples that doesn't belong to $c$th class. $N_{ij}^b = {1}/{N_c k_b}$ only if $x_j \in K^b(x_i)$ otherwise $N_{ij}^b=0$, and $N_{ij}^w = {1}/{N_c k_w}$ only if $x_j \in K^w(x_i)$ otherwise $N_{ij}^w=0$, where $N_c$ is the total number of elements in the $c^{th}$ class. The total-class dissimilarity matrix is obtained as $T^{spec} = S_b^{spec} + S_w^{spec}$. Finally the optimal projection matrix of LSC ($V$) is obtained by simultaneously maximizing the between-class dissimilarity matrix and minimizing the total-class dissimilarity matrix. 

\section{Proposed Spatial Spectral Approach}
Inspired by the existing literature \cite{zhou2015dimension} on spatial-spectral based DR, we propose to maximize the objective function of spatial-spectral based regularized graph local scaling cut method to achieve better performance instead of using spectral-based method alone.

\subsection{Spatial Guided Filters} \label{sec:SGF}
The guided filter \cite{kang2014spectral} is based on an assumption of local linear model, i.e., the filtering output $f$ is a linear transformation of the guidance image $I$ in a squared window $w_k$ of size $r \times r$ centered at the pixel $k$:
\begin{equation} \label{eq:gf_LT}
f_i = {a_k}{I_i} + {b_k} \,\,\,\,\, \forall i \in {w_k}
\end{equation}
where ${a_k}$ and ${b_k}$ are some linear coefficients in $w_k$. The assumption of this model ensures that ${\nabla}f \approx a{\nabla}I$, i.e., the filtering output $f$ has an edge if the guidance image $I$ has an edge at that location. These linear coefficients ${a_k}$ and ${b_k}$ are determined by minimizing the energy function:
\begin{equation} \label{eq:gf_energy}
E({a_k},{b_k}) = \sum_{i \in w_k}{(({a_k}{I_i} + {b_k}-P_i)^2+ \epsilon {a_k}^2)}
\end{equation}
Here, $P$ is the input image, and $\epsilon$ is the regularization parameter used to determine the degree of blurring for the guided filter. The solution of this energy function minimizes the difference between filtered image $f$ and input image $P$ by maintaining the linear model. 

In this filtering step, we obtain the first PCA coefficient of the input image, so that maximum reconstruction is possible. Then, we replicate the reconstructed image in spatial domain for each band, forming the tensor with dimensions same as data $P$. This filter captures different small and large homogeneous spatial structure of the HSI and preserves their edges.

\subsection{Spectral Regularized LSC (RLSC)}
To extract the spectral-domain information, we propose a regularized LSC (RLSC) method. The proposed RLSC method preserves the original distribution of the data and overcomes the singularity problem generated by pairwise dissimilarity matrix of the data samples.
The RLSC performs the spectral-domain local scaling cut operation. Inspired by the existing literatures on spectral domain DR methods in \cite{zhou2015dimension} and \cite{zhang2015scaling}, we propose a new objective function for the RLSC criteria. The objective function is defined as
\begin{equation}
\begin{split}
RS_b^{spec} = & tr(V^T [(1-\alpha)S_b^{spec} + \alpha XX^T] V); \\
RS_w^{spec} = & tr(V^T [(1-\alpha)S_w^{spec} + \alpha (\mbox{\textit{Diag}}(diag(S_w^{spec})))] V) \\
T^{spec} = & RS_b^{spec} + RS_w^{spec} \\
    = & tr(V^T[(1-\alpha)(S_w^{spec} + S_b^{spec}) + \alpha (R_w+R_b)]V) \\
RLSC(V) &= \mathop {\max }\limits_{V\, \in \,{R^{D \times d}}} \, \frac{RS_b^{spec}}{T^{spec}}\\        
 = \mathop {\max }\limits_{V\in {R^{D \times d}}} & \frac{tr(V^T [(1-\alpha)S_b^{spec} + \alpha R_b] V)}{tr(V^T[(1-\alpha)(S_w^{spec} + S_b^{spec}) + \alpha (R_{w}+{R_b})] V)}
\end{split}
\end{equation}
where $R_{w} = \mbox{\textit{Diag}}(diag(S_w^{spec})))$ and $R_{b} = XX^T$ are the regularizers of the within-class dissimilarity $S_w^{spec}$ and between-class dissimilarity $S_b^{spec}$ respectively. $\alpha \in [0,1]$ is the regularization parameter, $tr(\cdot)$ is the trace of a matrix, $diag(\cdot)$ represents a vector that contain the diagonal elements of a matrix, and $\mbox{\textit{Diag}}(\cdot)$ converts the vector into a diagonal matrix. The numerator ($RS_b^{spec}$) of the objective function corresponds to the between-class dissimilarity with the regularization term $R_b$ and the denominator ($T^{spec} = RS_b^{spec} + RS_w^{spec} $) represents the combination of within-class ($RS_w^{spec}$) and between-class ($RS_b^{spec}$) dissimilarity matrix with their corresponding regularizer $R_w$ and $R_b$.

The major contribution in this spectral part is the regularization terms $R_b$ and $R_w$. The covariance regularizer $R_b$ in the numerator corresponds to the variance of the data. This $R_b$ maximizes the between class data variance and preserve the data diversity \cite{zhou2015dimension}, \cite{weinberger2006introduction}. Whereas the diagonal regularizer $R_w$ improves the stability by decreasing the large eigenvalues and increasing small eigenvalues. This reduces the decay of the eigenvalues and retains the discriminative informations. This makes RLSC more stable.  
However, in the denominator, the total dissimilarity combines the both $RS_b^{spec}$ and $RS_w^{spec}$. Hence, the denominator provides both the diversity and stability to the solution.

When $\alpha = 0$ the RLSC becomes the LSC. The RLSC uses the labeled samples to determine the discriminative projection direction by considering the original distribution and modality. The regularization terms are added to the between-class and within-class dissimilarity matrices to incorporate the data diversity and avoid the singularity issue in the local manifold structure of the neighborhood samples.  

\subsection{Spatial Neighboring Pixel LSC (NPLSC)}
In spatial-domain, the neighboring pixels in the locality mostly belong to the same class or contains similar material. Hence, this spatial information can be useful in determining the projection matrix to improve the classification accuracy. In the proposed neighboring pixel local scaling cut (NPLSC) method, we construct a spatial-domain dissimilarity matrix using the spatial neighborhood pixel information of the filtered HSI data. It preserves the original spatial neighborhood pixel correlation in the projected NPLSC embedding space.  
Here, we first compute the number of spectral neighbors $K(x_i)$ of element $x_i$  by using the spectral K-nn. Next, we determine the spatial neighborhood of the pixel $x_j$ ($ x_j \in K(x_i)$).
Let the spatial patch of pixel $x_j$ be denoted by $P_j = \{x_{j1}, x_{j2}, ...,x_{jp}\}$ with $p$ surrounding pixels in the spatial neighborhood. Then, using the spatial patch elements of the spectral neighborhood, we compute both the between-class ($S_b^{spa}$) and within-class ($S_w^{spa}$) dissimilarity matrix
\begin{equation}
S_b^{spa} = \sum _{c=1}^{C}\sum _{x_i \in {U_c}} \sum _{x_j \in K^b(x_i)} \sum _{k=1}^p \eta_{ijk} (x_i - x_{jk})(x_i - x_{jk})^T 
\end{equation}
\begin{equation}
S_w^{spa} = \sum _{c=1}^{C}\sum _{x_i \in U_c} \sum _{x_j \in K^w(x_i)} \sum _{k=1}^p \eta_{ijk} (x_i - x_{jk})(x_i - x_{jk})^T
\end{equation}

where $\eta_{ijk} = \frac{w_{ijk}}{\sum _{t=1}^p{w_{ijt}}}$, and $w_{ijk} = \exp\{- \gamma ||(x_i - x_{jk})||^2\}$ is the weight based on the distance between the input and neighboring elements. $x_j$ is the central element and $x_{jk}$ is the $k$th spatial neighboring element of $x_j$. $N_c$ is the number of elements in the $c$th class. Here, $ K^b(x_i) \in \bar{U_c}$ and $K^w(x_i) \in U_c$ represent the between-class and within-class spectral neighbors. 

The NPLSC seeks linear projection matrix by minimizing the within-class dissimilarity matrix $S_w^{spa}$ and maximizing the between-class dissimilarity matrix $S_b^{spa}$.

\subsection{Spatial-Spectral Regularized LSC (SSRLSC)}
 
In HSI data, sometimes objects from different classes share similar spectral property; for example, concrete roads and roof tops share almost similar spectral signature although they belong to two different classes. Hence, only spectral distance measure is inadequate for determining optimal projection matrix. The spectral RLSC method exploits the local intrinsic manifold of the neighborhood data samples by using the localized $k$-nn graph.

On the contrary, NPLSC uses the spatial information to retain the local pixel neighborhood structure. However, NPLSC fails to connect two pixels with spatially higher pixel distance in a homogeneous region. In such a case, the labeled spectral information plays a vital role to establish a connection, which improves the discrimination criteria.

Hence, labeled spectral information and spatial information complements each other in terms of information content and thereby improves the HSI classification. We combine the information from both the domains and propose a spatial-spectral information based RLSC (SSRLSC) method. By combining the spectral based RLSC and spatial based NPLSC method, we construct spatial-spectral between-class dissimilarity matrix $SS_b$ and within-class dissimilarity matrix $SS_w$ as
\begin{align}
SS_w = \beta (RS_w^{spec}) &+ (1 - \beta)S_w^{spa} \\
SS_b = \beta (RS_b^{spec}) &+ (1 - \beta)S_b^{spa} \\
T_{ss} = SS_w &+ SS_b \\
SSRLSC(V) = & \mathop {\max }\limits_{V\, \in \,{R^{D \times d}}} \, \frac{SS_b}{T_{ss}}  \label{eq:SSRLSC_obj}      
\end{align} 

where $\beta \in [0,1]$ balances the contribution of the spectral and spatial contribution, and $T_{ss}$ is the spatial-spectral total class dissimilarity matrix.  The optimal projection matrix $V = [v_1, v_2, ..., v_d]$ of SSRLSC is obtained by solving by the generalized eigenvalue problem in (\ref{eq:SSRLSC_obj}).
The obtained projection matrix project the original data to a lower dimensional space spanned by $V$ to get the new feature vectors. 

\begin{table*}[!h]
\centering
\caption{The highest OA with its corresponding dimensionality (within the bracket), AA and $k$ for Botswana data(in \%)}
\label{tab:bots_all_sample}
\resizebox{0.92\textwidth}{!}{
\begin{tabular}{c|c|ccccccc|c|c|c|c|c}
\hline
\multirow{2}{*}{\begin{tabular}[c]{@{}c@{}}Train\\ Samples\end{tabular}} & \multirow{2}{*}{Methods} & \multirow{2}{*}{RAW} & \multirow{2}{*}{LFDA} & \multirow{2}{*}{NWFE} & \multirow{2}{*}{SELDLPP} & \multirow{2}{*}{SELDNPE} & \multirow{2}{*}{SSLSC} & \multirow{2}{*}{RLDE} & \multirow{2}{*}{SSRLDE} & \multicolumn{2}{c|}{\textbf{RLSC}} & \multicolumn{2}{c}{\textbf{SSRLSC}} \\ \cline{11-14} 
 &  &  &  &  &  &  &  &  & WMF & \textbf{Filter} & \multicolumn{1}{c|}{\begin{tabular}[c|]{@{}c@{}}\textbf{No} \\ \textbf{Filter}\end{tabular}} & \textbf{Filter} & \multicolumn{1}{c}{\begin{tabular}[c]{@{}c@{}}No \\ \textbf{Filter}\end{tabular}} \\ \cline{1-14} 
 \multirow{3}{*}{10} & OA & 77.76(145) & 78.59 (18) & 80.84 (14) & 75.08 (44) & 84.93 (38) & 80.71 (48) & 87.46 (40) & 87.54 (40) & 96.68 (48) & 86.76 (6) & 97.90 (48) & 88.28 (18) \\
 & AA & 79.87 & 79.34 & 82.43 & 77.56 & 85.99 & 82.35 & 88.39 & 88.02 & 94.35 & 87.25 & 97.83 & 89.12 \\
 & $k$ & 75.95 & 68.16 & 79.26 & 73.07 & 83.67 & 79.13 & 86.49 & 86.41 & 96.40 & 84.98 & 97.73 & 87.30 \\ \hline
\multirow{3}{*}{12} & OA & 79.77(145) & 79.88 (20) & 81.58 (20) & 78.81 (18) & 85.90 (38) & 81.77 (50) & 86.53 (50) & 87.64 (42) & 97.19 (44) & 86.84 (8) & 98.13 (50) & 89.80 (12) \\
 & AA & 81.97 & 80.95 & 82.74 & 80.12 & 86.53 & 83.15 & 87.46 & 88.16 & 97.15 & 88.01 & 98.17 & 90.59 \\
 & $k$ & 78.10 & 78.66 & 80.04 & 77.07 & 84.72 & 80.26 & 85.40 & 86.60 & 96.96 & 85.53 & 97.97 & 88.94 \\ \hline
\multirow{3}{*}{15} & OA & 80.37(145) & 80.85 (10) & 81.77 (10) & 78.24 (32) & 86.71 (36) & 83.19 (42) & 86.79 (48) & 87.09 (48) & 97.37 (50) & 87.27 (8) & 98.56 (32) & 90.80 (44) \\
 & AA & 82.47 & 81.06 & 83.33 & 80.22 & 87.83 & 84.57 & 87.68 & 87.44 & 97.94 & 88.26 & 98.63 & 91.49 \\
 & $k$ & 78.75 & 79.32 & 80.26 & 76.46 & 85.60 & 81.78 & 85.68 & 86.00 & 97.15 & 86.27 & 98.44 & 90.03 \\ \hline
\multirow{3}{*}{20} & OA & 81.91 (145) & 82.08 (12) & 83.09 (32) & 80.11 (30) & 87.05 (20) & 84.74 (42) & 87.08 (50) & 87.64 (32) & 98.21 (38) & 87.34 (10) & 99.33 (40) & 92.19 (26) \\
 & AA & 83.84 & 82.87 & 84.39 & 81.27 & 87.91 & 86.21 & 88.12 & 88.29 & 98.59 & 88.33 & 99.34 & 93.17 \\
 & $k$ & 80.41 & 80.08 & 81.67 & 78.46 & 85.95 & 83.46 & 85.99 & 86.60 & 98.06 & 86.59 & 99.27 & 91.53 \\ \hline
\multirow{3}{*}{25} & OA & 83.17 (145) & 84.43 (14) & 84,37 (20) & 81.29 (34) & 87.78 (30) & 86.05 (30) & 88.02 (40) & 88.53 (18) & 98.56 (48) & 88.10 (42) & 99.37 (46) & 93.33 (16) \\
 & AA & 85.20 & 85.29 & 85.79 & 83.08 & 88.61 & 87.27 & 89.09 & 88.86 & 98.52 & 89.24 & 99.37 & 94.08 \\
 & $k$ & 81.77 & 82.09 & 83.05 & 79.73 & 86.74 & 84.87 & 87.01 & 87.55 & 98.44 & 86.81 & 99.32 & 92.76 \\ \hline
\multirow{3}{*}{30} & OA & 84.25 (145) & 85.84 (38) & 85.11 (44) & 82.31 (22) & 88.07 (40) & 86.53 (24) & 88.39 (32) & 89.46 (32) & 98.42 (30) & 88.75 (15) & 99.55 (46) & 93.74 (32) \\
 & AA & 86.24 & 86.79 & 86.64 & 84.00 & 88.89 & 87.97 & 89.21 & 89.85 & 98.63 & 89.92 & 99.50 & 94.19 \\
 & $k$ & 82.93 & 82.54 & 83.85 & 80.83 & 87.05 & 85.38 & 87.40 & 87.45 & 98.29 & 87.62 & 99.52 & 93.21 \\ \cline{1-14} 
 \multicolumn{2}{c|}{ Time (N=10 \& D =30)} & 3.45 & 0.17 & 5.37 & 1.73 & 2.34 & 3.24 & 0.30 & 4.57 & 7.15 & 5.78 & 18.60 & 15.76 \\ \hline 
\end{tabular}}
\end{table*}


\begin{table*}[!h]
\centering
\caption{The highest OA with its corresponding dimensionality (within the bracket), AA and $k$ for Salinas data (in \%)}
\label{tab:salina_all_sample}
\resizebox{0.92\textwidth}{!}{
\begin{tabular}{c|c|ccccccc|c|c|c|c|c}
\hline
\multirow{2}{*}{\begin{tabular}[c]{@{}c@{}}Train\\ Samples\end{tabular}} & \multirow{2}{*}{Methods} & \multirow{2}{*}{RAW} & \multirow{2}{*}{LFDA} & \multirow{2}{*}{NWFE} & \multirow{2}{*}{SELDLPP} & \multirow{2}{*}{SELDNPE} & \multirow{2}{*}{SSLSC} & \multirow{2}{*}{RLDE} & \multirow{2}{*}{SSRLDE} & \multicolumn{2}{c|}{\textbf{RLSC}} & \multicolumn{2}{c}{\textbf{SSRLSC}} \\ \cline{11-14} 
 &  &  &  &  &  &  &  &  & WMF & \textbf{Filter} & \multicolumn{1}{c|}{\begin{tabular}[c|]{@{}c@{}}\textbf{No} \\ \textbf{Filter}\end{tabular}} & \textbf{Filter} & \multicolumn{1}{c}{\begin{tabular}[c]{@{}c@{}}No \\ \textbf{Filter}\end{tabular}} \\ \cline{1-14}
\multirow{3}{*}{10} & OA & 80.05 (204) & 83.15 (36) & 83.62 (44) & 76.94 (14) & 82.66 (46) & 74.88 (14) & 84.55 (16) & 86.20 (24) & 90.51 (36) & 84.94 (26) & 92.35 (40) & 86.73 (8) \\
 & AA & 86.78 & 89.63 & 89.26 & 81.75 & 89.84 & 81.33 & 90.37 & 90.98 & 91.30 & 90.48 & 95.78 & 92.28 \\
 & $k$ & 77.82 & 81.24 & 81.97 & 74.24 & 80.80 & 72.17 & 82.79 & 84.62 & 89.46 & 82.04 & 91.50 & 85.25 \\ \hline 
\multirow{3}{*}{12} & OA & 78.16 (204) & 84.36 (34) & 84.73 (22) & 75.92 (20) & 83.94 (50) & 76.11 (40) & 85.80 (14) & 87.30 (16) & 89.84 (46) & 85.16 (24) & 92.71 (22) & 87.69 (24) \\
 & AA & 86.14 & 90.57 & 90.61 & 82.94 & 89.76 & 80.68 & 91.08 & 91.62 & 91.04 & 91.04 & 95.82 & 93.18 \\
 & $k$ & 75.84 & 82.58 & 82.14 & 73.40 & 82.10 & 73.36 & 84.20 & 85.84 & 88.67 & 82.32 & 91.89 & 86.32 \\ \hline 
\multirow{3}{*}{15} & OA & 81.77 (204) & 84.50 (20) & 85.01 (24) & 77.95 (42) & 84.94 (46) & 77.57 (38) & 85.14 (48) & 87.14 (30) & 92.58 (36) & 86.45 (40) & 93.75 (30) & 88.48 (18) \\
 & AA & 89.05 & 91.13 & 91.76 & 84.35 & 90.85 & 83.38 & 91.09 & 91.63 & 92.83 & 91.28 & 96.38 & 93.42 \\
 & $k$ & 79.80 & 82.79 & 82.87 & 75.57 & 83.23 & 75.09 & 83.47 & 85.65 & 91.77 & 82.69 & 93.04 & 87.17 \\ \hline 
\multirow{3}{*}{20} & OA & 81.36 (204) & 86.28 (34) & 86.39 (26) & 79.51 (44) & 85.77 (16) & 79.24 (16) & 86.66 (16) & 87.99 (42) & 92.97 (38) & 87.06 (22) & 94.22 (36) & 89.00 (22) \\
 & AA & 89.48 & 92.15 & 92.13 & 84.82 & 91.29 & 84.82 & 91.63 & 92.07 & 93.80 & 91.72 & 96.82 & 94.26 \\
 & $k$ & 79.36 & 84.74 & 83.22 & 77.25 & 84.13 & 76.85 & 85.07 & 86.58 & 92.18 & 85.42 & 93.57 & 87.75 \\ \hline 
\multirow{3}{*}{25} & OA & 81.66(204) & 86.79 (32) & 86.25 (18) & 78.80 (38) & 86.42 (42) & 80.05 (20) & 86.69 (42) & 88.19 (42) & 92.87 (48) & 87.51 (28) & 94.45 (16) & 89.13 (16) \\
 & AA & 89.57 & 92.54 & 92.40 & 86.26 & 92.47 & 86.48 & 92.06 & 92.56 & 94.93 & 91.85 & 97.17 & 94.48 \\
 & $k$ & 79.69 & 85.30 & 83.78 & 76.56 & 84.93 & 77.80 & 85.17 & 86.85 & 92.07 & 85.59 & 93.83 & 87.91 \\ \hline 
\multirow{3}{*}{30} & OA & 82.48 (204) & 86.96 (48) & 86.92 (22) & 79.50 (36) & 86.85 (34) & 81.18 (50) & 86.69 (28) & 88.33 (22) & 93.38 (50) & 87.86 (22) & 95.21 (28) & 89.28 (10) \\
 & AA & 90.21 & 92.96 & 92.78 & 86.85 & 92.51 & 87.54 & 92.38 & 92.62 & 95.54 & 91.98 & 97.61 & 94.77 \\
 & $k$ & 80.57 & 85.49 & 85.55 & 77.32 & 85.38 & 78.98 & 85.21 & 87.01 & 92.62 & 85.73 & 94.67 & 88.06 \\ \hline 
  \multicolumn{2}{c|}{ Time (N=10 \& D=30)} & 42.60 & 0.25 & 5.91 & 1.85 & 2.75 & 4.75 & 0.39 & 10.10 & 7.83 & 5.27 & 27.01 & 24.32 \\ \hline 

\end{tabular}}
\end{table*}


\section{Experiments and Results}

We evaluated the performance of the proposed SSRLSC on two HSI datasets: Botswana $(N = 1476 \times 256, \, D = 145, \, C = 14)$, and Salinas $(N = 512 \times 217, \, D=204, \, C=16)$. Then, we compared it with state-of-the-art local fisher discriminant analysis (LFDA) \cite{sugiyama2007dimensionality}, nonparametric weighted feature extraction (NWFE) \cite{kuo2004nonparametric}, semi-supervised local discriminant analysis (SELDLPP \& SELDNPE) \cite{liao2013semisupervised}, semi-supervised local scaling cut (SSLSC) \cite{zhang2013semisupervised}, regularized local discriminant embedding (RLDE), and spatial-spectral regularized local discriminant embedding (SSRLDE) with weighted mean filter (WMF) \cite{zhou2015dimension}. The parameters used in these methods are selected based on their corresponding literature. The evaluation of SSRLSC is carried out by determining the overall classification accuracy (OA), class average accuracy (AA), and kappa coefficient ($k$)  of the linear support vector machine (SVM) classifier on the projected data. 
For this experiment, we select $10, 12, 15, 20, 25$ and $30$ random points from each class for training and remaining data points are used for testing. All the obtained results are the average of five iterations. 
\begin{table}[h!]
\centering
\caption{Variation of OA (\%) w.r.t. spatial window size w}
\label{tab:windows_OA}
\resizebox{0.49\textwidth}{!}{%
\begin{tabular}{c|llllllll}
\hline
w    & $1\times1$ & $3\times3$ & $5\times5$ & $7\times7$ & $9\times9$ & $11\times11$ & $13\times13$ & $15\times15$ \\ \hline
Botswana &     95.84       &  \textbf{ 97.72}       &     96.49       &     97.56       &     96.63       &      97.38        &  96.24            &      96.69        \\ 
Salina &     90.01       &    \textbf{91.87}        &    91.84        &    90.52        &    91.28        &    91.67          &   90.21           &  91.27            \\ \hline
\end{tabular}%
}
\end{table}
\subsection{Parameter Sensitivity}
In the proposed algorithm, several parameters are used, such as between-class ($k_b$) and within-class ($k_w$) neighborhood, spectral regularizer $\alpha$, spatial-spectral regularizer $\beta$, and spatial neighborhood window $w$. In the experiments, we considered $k_w  = k_b$ to avoid the data imbalance problem in computing the between-class and within-class dissimilarity matrices. We empirically set $k_w = k_b = 7$. 
Then $\alpha$ and $\beta$ parameters are tuned from the set $\{0.1,0.2,...,0.9\}$. After experimenting over the aforementioned range of values, we found that the maximum OA is achieved at parameter pair $(\alpha= 0.5)$ and $( \beta = 0.3)$ \footnote{The parameter analysis of $\alpha$ and $\beta$ is given in supplementary material.}. Note that we use these values of $\alpha$ and $\beta$ in rest of our experiments for both the datasets.  
The optimal value of $w$ is obtained empirically by varying the window size from $1 \times 1$ to $15 \times 15$ for training sample $10$ and dimensions $20$. As shown in Table~\ref{tab:windows_OA}, we found that the maximum OA is achieved at window size $3 \times 3$. The bigger size spatial window leads to higher probability of interference from pixel of other classes. Hence, we selected the window $w$ of size $3 \times 3$ or $p=9$ number of spatial neighbors to reduce the interference and it is used in subsequent experiments. 

\subsection{Performance Evaluation}
\begin{figure}[htp]
	\centering
	\begin{subfigure}{.24\textwidth}
		\centering
		\includegraphics[width=.99\linewidth]{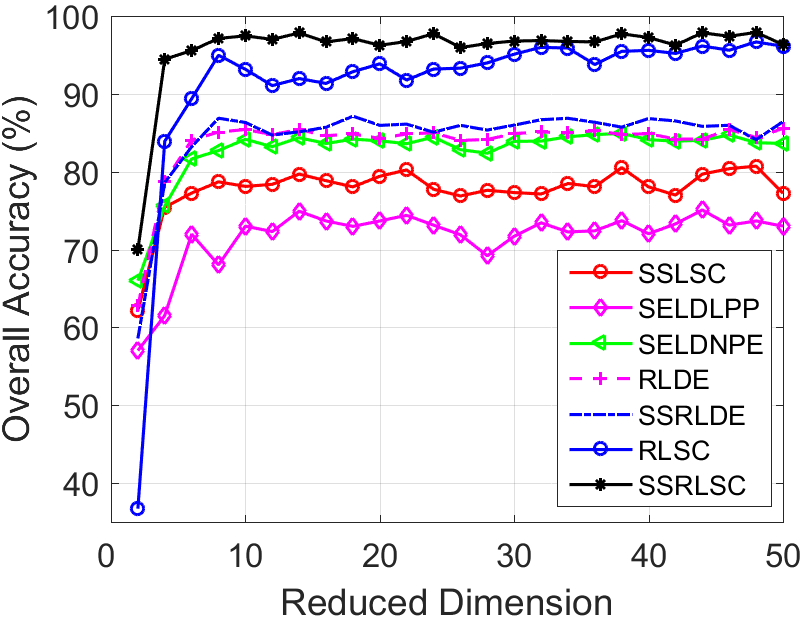}
		\caption{}
		\label{fig:All_Bots}
	\end{subfigure}    
  	\begin{subfigure}{.24\textwidth}
		\centering
		\includegraphics[width=.99\linewidth]{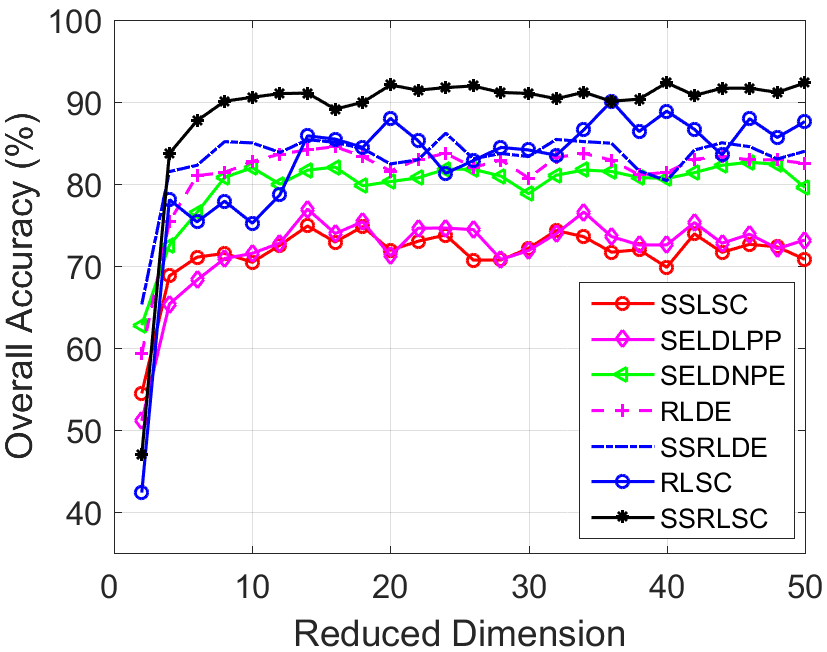}
		\caption{}
		\label{fig:All_Salina}
	\end{subfigure}
	\caption{ Effect of varying reduced dimensions on OAs for (\ref{fig:All_Bots})Botswana  and (\ref{fig:All_Salina})Salinas.}
	\label{fig:Test_RD_OAs}
\end{figure}

Fig.~\ref{fig:Test_RD_OAs} shows the variations of OAs with respect to the number of reduced dimensions when the train set consists of 10 samples per class. It can be observed that the proposed methods (SSRLSC and RLSC) significantly outperform other state-of-the-art DR methods. 
In both the datasets, SSRLSC achieves maximum accuracy even when the reduced dimension size is quite less.  Moreover, 
the performance of SSRLSC becomes stable at dimension $6$ and $8$ for Botswana (Fig.~\ref{fig:All_Bots}) and Salinas dataset (Fig.~\ref{fig:All_Salina}) respectively. 

The Table~\ref{tab:bots_all_sample} and \ref{tab:salina_all_sample} provide the statistics of the best OA and their corresponding measurements on Botswana and Salinas dataset respectively. 
We report the performances for different training sample size between $10-30$ and the reduced dimensionality is varied between $2 - 50$ and the best performance is provided in the table. As per Table~\ref{tab:bots_all_sample} and \ref{tab:salina_all_sample}, 
we can observe that, the proposed RLSC performs better than RLDE when the number of training samples used per class is more than $15$ for both the datasets. However, their performance is very close when less samples are used for training. When spatial information is incorporated without filter, SSRLSC also performs better than SSRLDE-WMF. In Botswana, we can observe that the performance margin is increased gradually with the use of more number of training samples. 
This signifies the effectiveness of the proposed DR algorithm for HSI datasets with fewer training samples. 
From Table~\ref{tab:bots_all_sample} and \ref{tab:salina_all_sample}, it is observed that the proposed the spectral RLSC running time is very competitive with the NWFE spectral method in Botswana dataset. However, it takes lesser time than NWFE in Salinas dataset. 

\subsection{Discussion}
The spatial filter preserves the edges and pixel consistency by performing edge-aware noise smoothening. This makes the filtered data consistent in homogeneous areas. Hence, the use of spatial neighboring pixels improves the discriminative ability of the projection directions, which can be observed in Table~\ref{tab:bots_all_sample} and \ref{tab:salina_all_sample}. 
The diagonal regularizers used in RLSC solve the singularity (stability) problem caused by small training samples. The diagonal regularizers solve this issue by reducing the decay of eigenvalues. This counters the decay by acting against the bias estimation of the small eigenvalues based on limited training samples \cite{friedman1989regularized}. Similarly, the covariance regularizer preserve the maximum data variance. The higher data variance enhances the data diversity \cite{zhou2015dimension} and \cite{weinberger2006introduction}. As per the results shown in Table~\ref{tab:bots_all_sample}, \ref{tab:salina_all_sample} and Figure~\ref{fig:Test_RD_OAs}, we can observe that the SSRLSC always performs better than RLSC. Hence, the spatial information of NPLSC acts as a performance booster on SSRLSC by complementing the labeled spectral informations in RLSC to achieve better projection matrix. The projection matrix in SSRLSC, not only preserves the spectral domain local euclidean neighborhood class relation but also spatial domain local neighboring pixel structures. 

\section{Conclusion}
This work proposed a novel spatial-spectral DR method, called SSRLSC, for HSI classification. This method combines two new methods, the spectral RLSC and spatial NPLSC. The guided filter used in this method increases the neighboring pixel consistency to preserve the spatial contextual information and discriminates the edges of the complimentary informations robustly. The regularization strategy in the spectral RLSC overcomes the data singularity by diversifying the HSI data samples. This enhances the discrimination capability and improves the classification accuracy. The NPLSC method is a robust graph cut based spatial segmentation technique, which incorporates the unlabeled spectral neighborhood measure with the spatial pixel neighborhood correlation to construct the spatial dissimilarity matrix for HSI data. 
The spatial NPLSC in SSRLSC solve the class identification and class variation problem caused due to sole use of spectral domain measurement.  
The promising experimental results on these two benchmark HSI data sets demonstrate the robustness and efficiency of the proposed DR algorithm.

\bibliographystyle{IEEEtran}
\bibliography{IEEEabrv,myBib}

\end{document}